\documentclass[a4]{article}
\usepackage{graphicx}
\usepackage{algorithm2e}
\usepackage{amssymb}
\usepackage{hyperref}

\def\be{\begin{equation}}
\def\ee{\end{equation}}
\newcommand{\ff}[1]{{\bf  #1}}
\def\x{\ff{x}}
\def\v{\ff{v}}

\makeindex

\begin{document}

\title{Why the Firefly Algorithm Works?}

\author{Xin-She Yang$^1$, Xing-shi He$^2$ \\[10pt]
1)  School of Science and Technology, Middlesex University, London NW4 4BT, UK.\\
2) College of Science, Xi'an Polytechnic University, Xi'an, China.
}

\date{}

\maketitle

\begin{center}
\parbox{5.2in}{
\noindent {\small {\bf Citation Detail}: X.-S. Yang, X.S. He, Why the firefly algorithm works?
in: {\it Nature-Inspired Algorithms and Applied Optimization} (Edited by X.-S. Yang), Springer,  pp. 245-259 (2018).
\href{https://doi.org/10.1007/978-3-319-67669-2_11}{https://doi.org/10.1007/978-3-319-67669-2\_11}
}  }
\end{center}

\abstract{Firefly algorithm is a nature-inspired optimization algorithm and there have been significant developments since its appearance about ten years ago. This chapter summarizes the latest developments about the firefly algorithm and its variants as well as their diverse applications. Future research directions are also highlighted.

}

\section{Introduction}

Nature-inspired computation has become a new paradigm in optimization, machine learning, data mining and computational intelligence with a diverse range of applications.
The essence of nature-inspired computing is the nature-inspired algorithms such as
genetic algorithm (GA) \cite{Holland}, particle swarm optimization (PSO) \cite{Kennedy} and firefy algorithm (FA) \cite{Yang}. Most nature-inspired algorithms use some characteristics of swarm intelligence \cite{Fisher}, and an overview of swarm intelligence to nature-inspired computation was recently carried out by Yang \cite{YangIEEE}.

Among nature-inspired algorithms, firefly algorithm (FA) was developed by Xin-She Yang in late 2007 and early 2008 \cite{Yang}, and it is almost ten years since its development. Significant developments have been made in the last few years, and thus this chapter intends to provide a state-of-the-art review of FA and its variants with an emphasis on the most recent studies.

Therefore, this chapter is organized as follows. Section 2 introduces the fundamentals of the firefly algorithm, and Section 3 explains why FA works well in practice. Section 4 highlights the main differences between FA and PSO, and Section 5 summarizes some of the recent variants of FA.
Section 6 reviews some of the diverse applications of FA and its variants. Finally, Section 7 concludes with discussion of future research directions.

\section{Firefly Algorithm}

The bioluminescence flashes of fireflies are an amazing sight in the summer sky in tropical and temperate regions. It is estimated that there are about 2000 species of fireflies and most species produce short, rhythmic flashes. Each species can have different flashing patterns and rhythms, and one of the main functions of such flashing light acts as a signaling system to communicate with other fireflies. The rate of flashing, intensity of the flashes and the amount of time between flashes form part of the signaling system \cite{Lewis}, and female fireflies respond to a male's unique flashing pattern. Some tropical fireflies can even synchronize their flashes, leading to self-organized behaviour.

As light intensity in the night sky decreases as the distance from the flashing source increases, the range of visibility can be typically a few hundred metres, depending on weather conditions. The attractiveness of a firefly is usually linked to the brightness of its flashes and the timing accuracy of its flashing patterns.

\subsection{The Standard Firefly Algorithm}

 Based on the above characteristics, Xin-She Yang developed the firefly algorithm (FA) \cite{Yang,YangFA}. Inside FA, the attractiveness of a firefly is determined by its brightness.
 Due to exponential decay of light absorption and inverse-square law of light variation with distance, a highly nonlinear term is used to simulate the variation of light intensity or attractiveness.

 In the FA, the main algorithmic equation for the position $\x_i$ (as a solution vector to a problem) is
\be  \x_i^{t+1} =\x_i^t + \beta_0 e^{-\gamma r^2_{ij} }
(\x_j^t-\x_i^t) + \alpha \; \ff{\epsilon}_i^t, \label{FA-equ-100} \ee
where $\alpha$ is a scaling factor controlling the step sizes of the random walks,
while $\gamma$ is a scale-dependent parameter controlling the visibility of the fireflies (and thus
search modes). In addition, $\beta_0$ is the attractiveness constant when the
distance between two fireflies is zero (i.e., $r_{ij}=0$). In the above equation,
the second term on the right-hand side (RHS) is the nonlinear attractiveness
 which varies with distance, while the third term is a randomization term and
 $\ff{\epsilon}_i^t$ means that the random number vectors should be drawn from
 a Gaussian distribution at each iteration.

This system is a nonlinear system, which may lead to rich characteristics
in terms of algorithmic behaviour.
Loosely speaking, FA belongs to the category of swarm intelligence (SI) based algorithms,
and all SI-based algorithms use some aspects of swarming intelligence \cite{Fisher}.

It is worth pointing out that the distance $r_{ij}$ between firefly $i$ and
firefly $j$ can be defined as their Cartesian distance. However, for some problems
such as the internet routing problems, this `distance' can be defined as time delay.
For certain combinatorial problems, it can be defined even as Hamming distance \cite{Osaba}.
In addition, since the brightness of a firefly is associated with the objective landscape
with its position as the indicator, the attractiveness of a firefly seen by others, depending on their relative positions and relative brightness. Thus, the beauty is in the eye of the beholder. Consequently, a pair comparison is needed for comparing all fireflies. The main steps of FA can be summarized as the pseudocode in Algorithm~\ref{Alg-1}.

\begin{algorithm}
\hrule
Initialize all the parameters ($\alpha, \beta, \gamma, n$)\;
Initialize randomly a population of $n$ firefies\;
Evaluate the fitness of the initial population at $\x_i$ by $f(\x_i)$ for $i=1,...,n$\;
\While{($t<$ MaxGeneration)}{
\For{All fireflies ($i=1:n$)}
{\For{All other fireflies ($j=1:n$) (inner loop)}{
\If{Firefly $j$ is better/brighter than $i$}
{Move firefly $i$ towards $j$ according to Eq.~(\ref{FA-equ-100})\;}}
Evaluate the new solution and accept the new solution if better\; }
Rank and update the best solution found so far\;
Update iteration counter $t \gets t+1$\;
Reduce $\alpha$ (randomness strength) by a factor\;
} \hrule
\caption{Firefly algorithm. \label{Alg-1}}
\end{algorithm}

Furthermore, $\alpha$ is a parameter controlling the strength of the randomness or perturbations in FA. The randomness should be gradually reduced to speed up the overall convergence. Therefore, we can use
\be \alpha=\alpha_0 \theta^t, \ee
where $\alpha_0$ is the initial value and $0<\theta<1$ is a reduction factor. In most cases, we can use $\theta=0.9$ to $0.99$, depending on the type of problems and the desired quality of solutions.

In fact, since FA is a nonlinear system, it has the ability to automatically
subdivide the whole swarm into multiple subswarms. This is because short-distance
attraction is stronger than long-distance attraction, and the division of
swarm is related to the mean range of attractiveness variations.
After division into multi-swarms, each subswarm can potentially swarm
around a local mode. Consequently, FA is naturally suitable for multimodal
optimization problems. Furthermore, there is no explicit use of the best solution $\ff{g}^*$,
thus selection is through the comparison of relative brightness
according to the rule of `beauty is in the eye of the beholder'.

\subsection{Special Cases of FA}

To gain more insight, let us analyze the FA system more carefully.
By looking at Eq.~(\ref{FA-equ-100}) closely, we can see that $\gamma$ is an important scaling parameter \cite{Yang,YangFA}.

\subsubsection{Case A: $\gamma=0$}

At one extreme, we can set $\gamma=0$, which means that there is no exponential decay and thus the visibility is very high. In this case, all fireflies can see each other in the whole domain and we
have \be \x_i^{t+1}=\x_i^t +\beta_0 (\x_j^t-\x_i^t) +\alpha \epsilon_i^t. \ee

\begin{itemize}

\item If $\gamma=0$, $\alpha=0$ and $\beta_0$ is fixed, then FA becomes a variant of differential evolution (DE) without crossover \cite{Storn,Yang2014}.
In this special case, if we replace $\x_j$ by the best solution in the group $\ff{g}^*$,
this reduced FA is equivalent to
a special case of accelerated particle swarm optimization (APSO) \cite{Yang,Yang2014}.

\item If $\beta_0=0$, FA is equivalent to the basic simulated annealing (SA) with $\alpha$ as the cooling schedule \cite{Yang2014}. In addition, if $\epsilon_i$ is further replaced by $\epsilon \x_i$, this special case is equivalent to the pitch adjustment of the harmony search (HS) algorithm.

\end{itemize}

Thus, it is clear that DE, APSO, SA and HS are special cases of the standard FA. In other words,
FA can be considered as a good combination of APSO, HS, SA and DE enhanced in a nonlinear system.
It is no surprise that FA can outperform these algorithms for many applications.

\subsubsection{Case B: $\gamma \gg 1$}

At the other extreme when $\gamma \gg 1$,  the visibility range is very short.
Fireflies are essentially flying in a dense fog and they cannot see each other clearly. Thus, each firefly flies independently and randomly. In fact, the exponential term $\exp[-\gamma r_{ij}^2]$
will decrease significantly if $\gamma r_{ij}^2=1$, which means that the radius $R$ or range of influence can be defined by
\be R=\frac{1}{\sqrt{\gamma}}. \ee
Therefore, a good value of $\gamma$ should be linked to the scale or limits of the design variables so that the fireflies within a range are visible to each other.

For a given objective landscape, if the average scale of the domain is $L$, then $\gamma$ can be estimated by
\be \gamma=\frac{1}{L^2}. \ee
If there is no prior knowledge about its possible scale, we can start with $\gamma=1$ for most problems, and then increase or decrease it when necessary. In theory, $\gamma \in [0, \infty)$,
but in practice, we can use $\gamma=O(1)$, which means that we can use $\gamma=0.001$ to $1000$ for most problems we may meet.

\subsection{Discrete FA}
The standard FA was designed to solve continuous optimization problems. In order to solve  discrete optimization problems, some discretization techniques should be used. For example, one way of converting a continuous variable $x$ to a binary one is to use the sigmoidal function
\be S(x)=\frac{1}{1+e^{-x}}, \ee
where $S \rightarrow 1$ for $x \rightarrow \infty$, while $S \rightarrow 0$
for $x \rightarrow -\infty$. However, this S-shaped function requires a large range to get a proper conversion. In practice, many researchers use an additional rule with a random threshold. A common technique is to use a random number $r \in [0,1]$. If $S>r$, then $S=1$, otherwise, $S=0$. Obviously, once we have $S \in \{0, 1\}$, we can use $u=2S-1$ to get $u \in \{+1, -1\}$ if needed.

Another way of conversion is to use random permutation. For example, a set of a uniformly distributed random number such as  $r=[0.3,\; 0.9, ..., \;0.7]$ can be converted to integers.
On the other hand, an interesting conversion technique is to use a modulus function by
\be u = \lfloor x +k \rfloor \;\; \textrm{mod} \;\; m, \ee
to convert $x$ to an integer $u$. Here, $k$ and $m>0$ are integers.

There are other methods for discretization, including random keys, random permutation, Hamming-distance based method, tanh(x), and others \cite{Rodrig}.

Many studies using FA have demonstrated how the algorithm works and the effectiveness of the algorithm. Interested readers can refer to the book by Yang \cite{Yang2014} and reviews \cite{Fister,Tila2}.  Now let us explain in more detail why the algorithm works.

\section{Why the Firefly Algorithm Works?}

In the above descriptions, we have explained the main steps of the FA and how it works.
We now try to summarize why it works so well in practice.

The exact reasons why FA works may require further mathematical analysis, specially for the variants to be introduced later. As it still needs a theoretical framework to explain the working mechanisms  of FA and all other algorithms, we do not intend to figure out all the reasons why an algorithm works. However, from both empirical observations and the analysis of the algorithm structure, we can summarize the following four reasons why the FA works \cite{Yang2014}:

\begin{itemize}
\item From the special cases discussed in the previous section, we know that APSO, SA, HS and DE are special cases of FA, and thus FA can be considered as a good combination of all these algorithms. Therefore, it is no surprise that FA can work more efficiently than these algorithms.

\item Due to the nonlinear attraction mechanism in FA, the short-distance attraction is stronger than long-distance attraction; therefore, the whole swarm can automatically subdivide into multiple subswarms. Each swarm can potentially swarm around a local mode, and among all the local modes, there is always a global optimal solution. Consequently, the multiswarm nature of FA enables FA to find multiple optimal solutions simultaneously and FA is naturally suitable for solving nonlinear, multimodal optimization problems.
    Therefore, for a given problem with $m$ modes, if the number of fireflies $n$ is much higher than $m$ (i.e., $n \gg m$), then all the optima (including the global best) can be found simultaneously.

\item The influence radius or range is controlled by $\gamma$. As a small value of $\gamma$ means higher influence and higher visibility, while a higher value of $\gamma$ reduces its influence and visibility. Therefore, we can tune $\gamma$ to control the subdivision of the swarm. If $\gamma=0$, there is no subdivision and all fireflies belong to a single swarm.
    A moderate value of $\gamma$ leads to multiswarms, while a much higher value of $\gamma$ may lead to individual random walks without a swarm. As a result, the diversity and properties of the population are linked to $\gamma$. This nonlinearity provides much richer dynamic characteristics.

\item In comparison with PSO and other algorithms, FA does not use velocities explicitly, which means that FA does not have any drawbacks associated with velocities. In addition, FA does not use $\ff{g}^*$ in its equation. The use of $\ff{g}^*$ can potentially lead to premature convergence if the initial $\ff{g}^*$ lies in the wrong region, which will attract all other agents towards it. Therefore, FA can avoid any disadvantage associated with $\ff{g}^*$.

\end{itemize}

It is worth pointing out that all these parameters have to be tuned properly. For example, $\alpha$ as the strength of the random walks must be reduced gradually; otherwise, the convergence may be slowed down by too much randomness. Similarly, a proper value of $\gamma$ has to be tuned to allow a good set of subswarms to emerge automatically \cite{Yang, YangCSFA}.

\section{FA is not PSO}

Though FA and PSO are both swarm intelligence based algorithms, they thus share some similarity;  however, FA is not PSO because they have some significant differences.
Apart from the different inspiration from nature, we briefly summarize here the main differences between FA and PSO:

\begin{itemize}
\item FA is a nonlinear system due to the nonlinear attraction term $\beta_0 \exp(-\gamma r_{ij}^2)$, while PSO is a linear system because its updating equations are linear in terms of $\x_i$ and $\v_i$. The nonlinear dynamic nature of FA can lead to much richer characteristics in terms of algorithmic behaviour and population properties.

\item The strong nonlinearity of FA means that FA has an ability of multi-swarming, while PSO cannot. Thus,  FA can find multiple optimal solutions simultaneously and consequently deal with multimodal problems more effectively.

\item PSO uses velocities, but FA does not. Thus, FA does not have the drawbacks associated with velocity initialization and instability for high velocities of particles.

\item FA has some scaling control (via $\gamma$), while PSO has no scaling control. Such scaling control can give FA more flexibility.

\end{itemize}

All these differences enable FA to search the design spaces more effectively for multimodal objective landscapes.

\section{Variants of FA}

Since the development of FA in 2008, it has been applied to many applications \cite{YangCSFA}.
A comprehensive review was done by Fister et al. in 2013 \cite{Fister}, covering the literature up to 2013. Yang and He provided another review from a different perspective in 2013 \cite{YangHe}. More recently, Tilahun et al. provided an updated review on the continuous versions of the firefly algorithm and its variants \cite{Tila2}, and the discrete versions of the firefly algorithms were also reviewed by Tilahun and Ngnotchouye in 2017 \cite{Tila}.

Despite the success of the standard FA, many variants have been developed to enhance its performance in the last few years. Again many of these variants have been reviewed by Fister et al. \cite{Fister} and Tilahun et al. \cite{Tila2}, and we will not repeat their coverage. Instead, here we will focus only on the most recent variants that have just appeared  in the last few years.

Though there are a diverse range of variants of FA, they can be loosely put into the following
six major variants/categories:
\begin{itemize}

\item {\bf Discrete FA}: The standard FA was designed to solve problems in the continuous domains. To solve discrete or combinatorial optimization problems, some modifications are needed.
For example, Marichelvam et al. developed a discrete FA for solving hybrid flow shop scheduling problems \cite{Marich,Marich2}, while Osaba developed a discrete FA for solving vehicle routing problems with recycling policy \cite{Osaba}. In addition, Poursalehi et al. used an effective discrete FA for optimizing fuel reload design of nuclear reactors \cite{Poursa}.
Zhang et al. used a discrete double-population for assembly sequence planning \cite{ZhangZ}.
These variants can be used to solve scheduling and planning problems as well as routing problems.

\item {\bf Adaptive FA}: In the standard FA, parameters are fixed, and it may be advantageous to use adaptively varying parameter values. Baykasoglu and Ozsoydan developed an adaptive firefly algorithm with chaos to solve mechanical design problems \cite{Bayk}, and G\'alvex and Iglesias \cite{Galvez} developed a memetic self-adaptive FA for shape fitting \cite{Galvez}.

\item {\bf Modified/Enhanced FA}: Researchers have designed various ways to modify and enhance the performance of FA. For example, Cheung et al. developed a non-homogeneous FA \cite{Cheung}, while Chou and Ngo developed a modified FA for multidimensional structural optimization \cite{Chou}. Darwish combined FA with a Bayesian classifer for solving classification problems \cite{Darwish}, while Fister et al. used quaternion to represent the solutions of FA in higher dimensions \cite{Fister2}.

    In addition, He and Huang used a modified FA for multilevel thresholding of color image segmentation \cite{HeLF}. Gupta used a modifed FA for controller design \cite{Gupta}. Tesch and Kaczorowska used a rotational FA for arterial cannula shape optimization \cite{Tesch}. Verma et al. developed an opposition and dimensional based modified firefly algorithm \cite{Verma}. Furthermore, Wang et al. developed a modified FA based on light intensity difference \cite{WangB}, while Wang et al. modified FA with neighborhood attraction \cite{WangH} and Yu et al. developed a variable step size FA \cite{Yu}.

    Additionally, Zhou et al. used an information-fusing FA for wireless sensor placement for structural monitoring \cite{Zhou}, and Zhou et al. combined FA with Newton's method to identify boundary conditions for transient heat conduction problems \cite{ZhouHL}.

\item {\bf Chaotic FA}: Some of the parameters in the FA can be replaced by the outputs of some chaotic maps, which may be able to enhance the exploration ability of the FA. For example,
Gandomi et al. developed a chaotic FA in 2013 \cite{Gandomi}, while Gokhale and Kale used a tent map for their chaotic FA \cite{Gokhale}. Also, Zouache et al. developed a quantum-inspired FA for discrete optimization problems \cite{Zou}, and Dhal et al. developed a chaotic FA for enhancing image contrast \cite{Dhal}.  Chaos-based FA variants were reviewed by Fister et al. \cite{Fister3}.

\item {\bf Hybrid FA}: Hybridization can be a good way to create new algorithm tools by combing
the advantages of each algorithm involved in the hybrid. For example, Aleshab and Abdullah developed a hybrid FA with a probabilistic neural network for solving classification problems \cite{Alwesh}, and Zhang et al. developed a hybrid by combing FA with DE and achieved
improved performance and accuracy \cite{ZhangLina}.

\item {\bf Multiobjective FA}: The standard FA was for single objective optimization
and Yang extended the standard FA to multiobjective firefly algorithm (MOFA) for design optimization \cite{YangMOFA}. In addition, Eswari and Nickolas developed a modified multiobjective FA for task scheduling \cite{Eswari}, while Wang et al. developed a hybrid multiobjective FA for big data optimization \cite{WangH2}, and Zhao et al. developed a decomposition-based multiobjective FA for RFID network planning with uncertainty \cite{Zhao}.

\end{itemize}

\section{Applications of FA and its Variants}

The applications of FA and its variants are diverse, a quick Google scholar search gives
more than 7000 outputs, and it is not possible to cover all these applications here.
It is not our intention to review even a good fraction of the applications in the current literature. For comprehensive reviews, interested readers can refer to \cite{Fister,Tila,Tila2}.  Here, our emphasis will be on the recent, new applications that can be representative in areas from engineering design to energy systems and from scheduling to image processing. For example, FA has been applied in the design of radial expanders in organic Rankine cycles \cite{Bahad},  design optimization of steel frames \cite{Carbas}, distributed generation system \cite{Chaur}, beam design \cite{Erdal}, wavelet neural network optimization \cite{Zainu}, hysteresis model identification \cite{Zaman}, detection of TEC seismo-ionospheric anomalies \cite{Akhoon} and structural search in chemistry \cite{Avenda}.

In the area of clustering and classification, Senthinath et al. compared and evaluated the performance of clustering using FA \cite{Senthin}. Gope et al. used FA for rescheduling of real power for congestion management concerning pumped storage hydro-units \cite{Gope}.
Long et al. used FA for heart disease predictions \cite{Long}.

For applications in design and optimization, Mohanty applied FA for designing shell and tube heat exchangers \cite{Mohanty}, while Shukla and Singh used FA to select parameters for advanced machining processes \cite{Shukla}. Hung applied FA in OFDM systems \cite{Hung} and
Kamarian et al. used FA for thermal buckling optimization of composite plates \cite{Kamar}, while
Jafari and Akbari used FA to optimize micrometre-scale resonator modulators \cite{Jafa}, and
Othman et al. used a supervised FA to achieve optimal placement of distributed generators \cite{Othman}. Also, Singh et al. combined FA with least-squares method
to estimate power system harmonics \cite{Singh}, and Kaur and Ghosh used a fuzzy FA for network reconfiguration of unbalanced distribution networks \cite{Kaur}.

In the area of energy engineering and energy systems, Ghorbani et al. used FA for prediction of gas flow rates from gas condensate reservoirs \cite{GhorbH}, and Massan et al. used FA to solve wind turbine applications \cite{Massan}. Wang et al. used an FA-BP neural network to forecast electricity price \cite{WangD} and Rastgou and Moshtagh used FA for multi-stage transmission expansion planning \cite{Rastgou}. In addition,  Satapathy et al. used a hybrid HS-FA based approach to improve the stability of PV-BESS diesel generator-based microgrid \cite{Sata}.

For image processing,  Kanimozhi and Latha used FA for region-based image retrieval \cite{Kani},
and Rajinikanth and Couceiro used an FA-based approach for color image segmentation \cite{Raji}. S\'achez et al. used FA to optimize modular granular
neural networks for human recognition \cite{Sanchez}.
Rahebi and Hardalac used FA for optic disc detection in retinal images \cite{Rahebi}, while
Gao et al. used FA for visual tracking \cite{Gao}, and Zhang et al. used a discrete FA for endmember extraction from hyperspectral images \cite{Zhang}.

In the area of time series and forecasting,
Xiao et al. used a combined model for electrical load forecasting \cite{Xiao} and
Ghorbani et al. used FA for capacity prediction in combination with support vector machine \cite{Ghorb}. In addition, Ibrahim and Khatib used a hybrid model for solar radiation prediction \cite{Ibrah}.

In the area of planning and navigation, Ma et al. used FA for planning navigation paths \cite{Ma}, and Patle et al. used FA to optimize mobile robot navigation \cite{Patle}.

In deep learning and software engineering, Rosa et al. used FA for learning parameters in deep belief networks \cite{Rosa}. Srivatsava used an FA-based approach for generating optimal software test sequences \cite{Sriva}, and  Kaushik et al. integrated FA in artificial neural network for predicting software costs accurately \cite{Kaush}.

Other applications include nanoscale structural optimization by Kougianos and Mohanty \cite{Kougi}, protein complex identification  by Lei et al. \cite{Lei},  protein structure prediction by Maher et al. \cite{Maher}. Also,
Nekouie and Yaghoobi used FA to carry out multimodal optimization \cite{Nekou}, and
Sundari et al. used an improved FA for programmed PWM in multilevel inverters with adjustable DC sources \cite{Sundari}.

\section{Conclusions and Future Directions}

As we have seen from the above reviews and discussions, FA and its variants have been successfully
applied in a wide spectrum of real-world applications. Despite its success, there are still some interesting future research directions concerning FA, and we will summarize them as follows.

\begin{enumerate}

\item {\it Theory}: Though we know FA and its variants work well, we do not have solid theoretical proof why they work and under exactly what conditions. A recent study by He et al. proved the global convergence of the flower pollination algorithm \cite{He2017}. It can be expected that the same methodology can be applied to analyze the firefly algorithm and other algorithms. Therefore, more theoretical analysis is needed.

\item {\it Adaptivity}: All bio-inspired algorithms including FA have parameters, and tuning of these parameters can be tedious. Ideally, algorithms should be able to tune their parameters using a self-tuning framework \cite{YangSTA} and also adapt their values to suit for a given type of problems. Future work can focus on the parameter adaptivity of FA and its variants.

\item {\it Hybrid}: Though there are many different variants of FA, it is no doubt that more hybrid variants will appear in the future. At the moment, hybridization is by trial and error, and it is not clear yet how to achieve a better hybrid by combining different algorithms, which needs more research and further insight.

\item {\it Co-evolution}: Simple hybridization can often work well; however, co-evolution can be more advantageous by co-evolving two or more algorithms together so as to allow the successful characteristics of an algorithm to enhance the co-evolutionary algorithm structure. It is not clear how to carry out co-evolution of algorithms.

\item {\it Applications}: In addition to the diverse range of applications reviewed in this chapter, there are more research opportunities of applying FA and its variants. Future applications can focus on the area in big data, deep learning and large-scale problems.
    Big data in combination with machine learning techniques such as deep nets can be an active research area for many years to come, and the nature-inspired algorithms can expect to play an important role in this area.

\end{enumerate}

As we can see that FA and its variants have been very successful in many applications, there are more opportunities for future research and applications. The authors hope that this work can inspire future research in the above mentioned directions with more real-world applications.


\end{document}